\documentclass{article}
\usepackage[preprint]{neurips_2026}
\bibpunct{[}{]}{,}{a}{,}{,}
\usepackage[utf8]{inputenc}
\usepackage[T1]{fontenc}
\usepackage{url,booktabs,amsmath,amssymb,amsfonts,microtype,graphicx,xcolor,float,array,tabularx,courier}
\usepackage{multirow}
\floatstyle{ruled}
\newfloat{algorithm}{tbp}{loa}
\floatname{algorithm}{Algorithm}
\usepackage{hyperref}
\usepackage{needspace}
\hypersetup{colorlinks=true,linkcolor=blue!50!black,citecolor=blue!50!black,urlcolor=blue!50!black,
 pdftitle={VACE: Validation-Gated Alternating Co-Evolution of Agent Models and Harnesses},
 pdfauthor={Jiexing Qi, Yu He, Jun Liu, Qichen Huang, Shaohua Hu, Zhan Dang, Guohua Chen, Rui Yang, Wen Jiang, Yang Liu, Tao Lyu, Fangming Li}}
\newcommand{\method}{VACE}
\newcommand{\methodfull}{Validation-Gated Alternating Co-Evolution of Agent Models and Harnesses}
\newcommand{\Rval}{\widehat R_{\mathcal V}}
\title{\method{}: \methodfull}
\author{%
  \textbf{Jiexing Qi\textsuperscript{1,*}, Yu He\textsuperscript{1}, Jun Liu\textsuperscript{1}, Qichen Huang\textsuperscript{1}, Shaohua Hu\textsuperscript{1,2,\textdagger}, Zhan Dang\textsuperscript{1}}\\
  \textbf{Guohua Chen\textsuperscript{1}, Rui Yang\textsuperscript{1}, Wen Jiang\textsuperscript{1}, Yang Liu\textsuperscript{1}, Tao Lyu\textsuperscript{1}, Fangming Li\textsuperscript{1,*}}\\[5pt]
  \textsuperscript{1}ICT AI Competence Center, Huawei Technologies Co., Ltd., Shanghai, China\\
  \textsuperscript{2}Shanghai Jiao Tong University, Shanghai, China\\
  \texttt{\{qijiexing, lifangming1\}@huawei.com}\\[3pt]
  {\small \textsuperscript{*}Corresponding authors: Jiexing Qi, Fangming Li}
}
\begin{document}
\raggedbottom
\maketitle
\begingroup
\renewcommand{\thefootnote}{\fnsymbol{footnote}}
\footnotetext[2]{Work done during internship at Huawei Technologies Co., Ltd.}
\endgroup

\begin{abstract}
Language model agents can be improved by updating their model weights or refining the harness that guides task execution. These components are coupled: weight updates change how the model uses the harness, while harness updates change the trajectories used for training. We propose \method{}, \textbf{V}alidation-Gated \textbf{A}lternating \textbf{C}o-\textbf{E}volution, which alternates agentic reinforcement learning with trajectory-driven harness refinement. After each RL stage, \method{} reuses the collected trajectories to propose a harness revision and evaluates the incumbent and candidate with the updated model held fixed. The candidate guides subsequent training only if it improves validation performance. With Qwen3.5-9B, \method{} achieves 45.26\% test accuracy on OfficeQA and a mean partial-credit score of 75.19\% on AutomationBench, exceeding weight-only RL by 6.43 and 9.09 percentage points and ungated alternation by 4.59 and 6.95 points, respectively. Across 44 harness proposals, 17 reduce validation performance at the updated checkpoint and are rejected before subsequent RL training, highlighting the importance of validation gating.
\end{abstract}
\section{Introduction}
\label{sec:intro}

Language-model agents solve tasks by reasoning over observations, invoking tools, and responding to feedback from their environments. Their performance depends on both the model and the system around it. Model weights determine how the agent interprets a task and chooses its actions, while instructions, reusable skills, tool interfaces, and execution rules determine how those capabilities are used. We refer to this surrounding system as the \emph{harness}.

Two complementary approaches improve these components. Agentic reinforcement learning (RL) trains model weights from task outcomes, with systems such as Agent Lightning~\citep{agentlightning} connecting agent execution to weight optimization. Harness optimization instead refines the instructions or procedures used by a model, for example through contextual playbooks~\citep{ace} or trajectory-guided revisions~\citep{harnessevolve}. Each approach can improve an agent, but optimizing one component while fixing the other leaves their interaction unaddressed. A skill designed for an early checkpoint may become unnecessarily restrictive after training, while an improved workflow may expose decisions that the model has not yet learned to make.

This interaction suggests improving the model and harness together. Consider a document-grounded agent whose skill instructs it to answer after a single retrieval. RL can improve reasoning over the retrieved evidence, but the skill may still lead the agent to stop before collecting enough information. Revising the skill to gather and compare additional evidence changes the executions from which the model learns. After further training, those executions reveal new opportunities for refinement. The connection is therefore bidirectional: model training supplies feedback for harness optimization, and the revised harness shapes the next stage of model training.

Harness refinement proposes discrete changes to instructions and skills from trajectory feedback, but these changes need not improve task performance with the updated model. For example, additional retrieval may improve evidence coverage while consuming the interaction budget on simpler tasks. An adopted revision also changes the trajectories used for subsequent RL, so its effect should be evaluated before it guides further training.

These considerations motivate \method{}, \emph{Validation-Gated Alternating Co-Evolution of Agent Models and Harnesses}, which combines alternating model--harness optimization with validation-gated harness updates. Each round trains the model under the current harness and reuses the collected trajectories to propose a revision. With the updated model held fixed, we compare the incumbent and candidate harnesses on the same validation tasks and adopt the candidate only if its validation score improves. The retained harness guides the next RL stage. We implement this loop through Uni-Agent~\citep{uniagent_github}, connecting agentic RL with trajectory-driven harness refinement. Here, \emph{co-evolution} refers to the repeated mutual adaptation of model weights and harness configurations.

\begin{figure}[t]
\centering
\includegraphics[width=.87\linewidth]{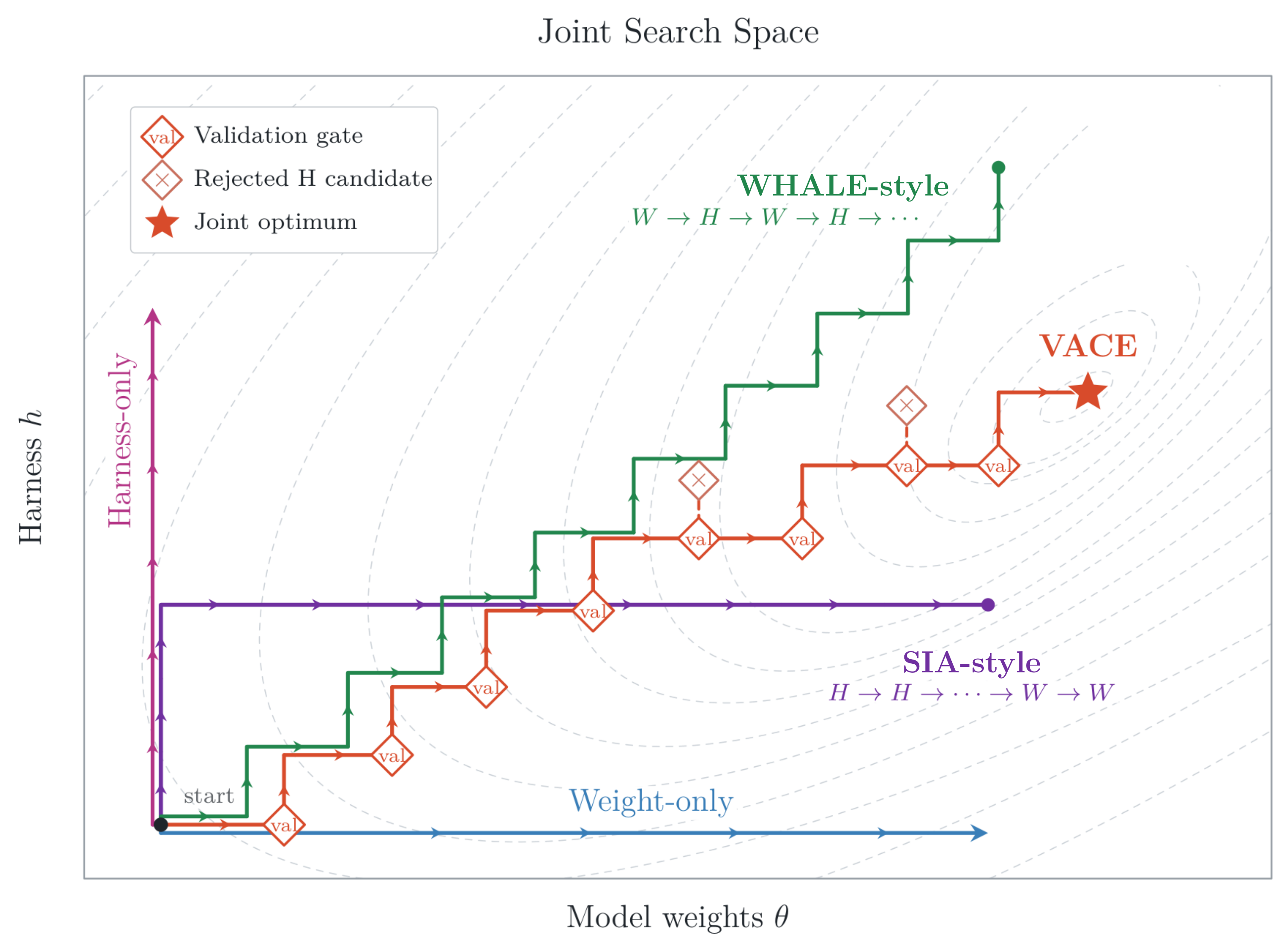}
\caption{Conceptual schedules for model--harness co-optimization. Harness-only and weight-only update one component. SIA-style and WHALE-style illustrate stagewise and alternating schedules, respectively; \method{} validates proposed harness updates before the next training round. The figure uses $\theta,h$ for the weights $W$ and harness $H$ in our notation. The paths and contours are schematic, and the star marks a conceptual joint optimum.}
\label{fig:joint_search_space}
\end{figure}

Model--harness co-optimization is an emerging research direction, explored in related systems such as SIA~\citep{sia} and Co-Harness~\citep{coharness}, and independently in concurrent work WHALE~\citep{whale}. \method{} contributes an RL-based co-evolution framework that connects trajectory feedback, repeated harness refinement, and checkpoint-specific validation in one training process. Figure~\ref{fig:joint_search_space} illustrates the update schedules, and Figure~\ref{fig:framework} details the proposed framework. Section~\ref{sec:related} discusses the technical relationships to related and concurrent work.

In our experiments on OfficeQA and AutomationBench with Qwen3.5-9B, \method{} scores 6.43 and 9.09 percentage points above weight-only RL, respectively, and 4.59 and 6.95 points above WHALE-style ungated alternation. \method{} also achieves a higher mean test score than SIA-style harness-first optimization on AutomationBench. Analysis of individual rounds reveals that 17 of 44 harness proposals reduce validation performance at the current checkpoint, illustrating why refinement and acceptance should be separate decisions.

\begin{samepage}
Our contributions are threefold:
\begin{itemize}
\item We propose \method{}, a framework for jointly improving agent models and harnesses through a closed loop of agentic RL and trajectory-driven harness refinement.
\item We introduce a checkpoint-specific validation mechanism within this loop, testing each proposed harness with the updated model before it shapes the next round of training.
\item We compare joint adaptation with single-component optimization and scheduling variants implemented with the same weight and harness optimizers on two agent benchmarks, and analyze how accepted and rejected revisions shape the co-evolution process.
\end{itemize}
\end{samepage}
\section{Related work}
\label{sec:related}

\paragraph{Model training.}
Agentic RL improves interactive reasoning and tool use~\citep{ragen,toolrl}. Search-R1~\citep{searchr1} trains models to interleave reasoning with search through RL, while WebAgent-R1~\citep{webagentr1} trains web agents through multi-turn environment interactions. AgentCPM-Explore~\citep{agentcpmexplore} studies long-horizon exploration with a compact agent model, combining model fusion, reward denoising, and context refinement. Agent Lightning~\citep{agentlightning} separates agent execution from RL training, enabling weight updates across different implementations. These approaches learn from execution feedback; \method{} also uses it to refine the harness.

\paragraph{Harness optimization.}
Reflexion~\citep{reflexion} stores verbal reflections from task feedback, and ExpeL~\citep{expel} extracts reusable insights from agent experience without parameter updates. GEPA~\citep{gepa} uses trajectory-based reflection to propose and evaluate prompt revisions. Agentic Context Engineering~\citep{ace} improves contextual playbooks, while the Darwin G\"odel Machine~\citep{dgm} evaluates changes to agent code. HarnessEvolve~\citep{harnessevolve} uses trajectory diagnostics and reference-guided error analysis to propose harness revisions, with quality checks and performance-based selection. Related approaches optimize prompts and modular pipelines~\citep{dspy,textgrad}. \method{} couples HarnessEvolve with model training.

\paragraph{Related and concurrent co-optimization frameworks.}
SIA~\citep{sia} uses a feedback agent to update both the harness and model weights; its reported experiments improve the harness before performing weight updates. Co-Harness~\citep{coharness} alternates failure-driven harness refinement with supervised training on successful trajectories and validates proposed harness changes. Developed concurrently with our work, WHALE~\citep{whale} alternates online rejection-sampling fine-tuning with executable-harness search and studies both fixed and adaptive phase durations. Its Meta-Harness search selects candidates through internal evaluations. \method{} combines agentic RL with trajectory-driven harness proposals, validating each proposal at the updated checkpoint before it guides subsequent training. We study this design on document-grounded reasoning and workplace workflows, with harness edits focused on skills and execution guidance.
\section{VACE}
\label{sec:method}

\begin{figure}[t]
\centering
\includegraphics[width=\linewidth]{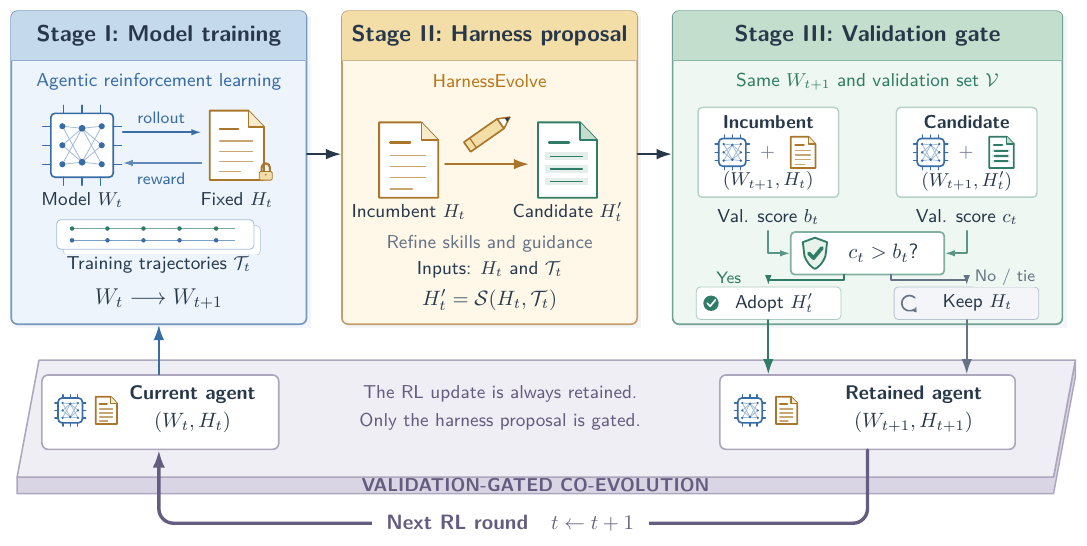}
\caption{Overview of \method{}. Stage I trains the model under the current harness and collects execution trajectories. Stage II turns this feedback into a proposed harness revision. Stage III compares the incumbent and candidate at the updated checkpoint. The retained harness then guides the next RL round, closing the feedback loop between model training and harness refinement.}
\label{fig:framework}
\end{figure}

\method{} jointly improves a model and its harness by connecting their updates through execution feedback. Figure~\ref{fig:framework} gives an overview. Each round first trains the model with the current harness, then uses the collected trajectories to propose a harness revision. Finally, the updated model executes validation tasks under both harnesses, and the better-scoring harness is retained, with ties favoring the incumbent. The retained pair starts the next round.

\Needspace{8\baselineskip}
\subsection{Model--harness co-optimization}
An agent state is a pair $(W,H)$, where $W$ denotes model weights and $H$ denotes the harness configuration. For a task $x$, execution produces a trajectory $\tau\sim p_{W,H}(\cdot\mid x)$ containing model decisions, tool calls, environment responses, and an outcome. Given a task distribution $\mathcal P$ and reward $r(x,\tau)$, the objective is
\begin{equation}
J(W,H)=\mathbb E_{x\sim\mathcal P}\mathbb E_{\tau\sim p_{W,H}(\cdot\mid x)}[r(x,\tau)].
\label{eq:objective}
\end{equation}
The objective $J(W,H)$ is the expected task reward. We maximize it by alternating updates to $W$ and $H$. Their coupling appears in $p_{W,H}$: a weight update changes how the model uses the harness, and a harness update changes the executions available for subsequent learning. Our implementation focuses harness edits on skills and execution guidance, while keeping the task evaluator, tool semantics, and environment transition rules fixed.

We distinguish training tasks $\mathcal D_{\mathrm{train}}$, validation tasks $\mathcal V$, and an independent test set. For a complete validation run, the score aggregates task outcomes as
\begin{equation}
\Rval(W,H)=\frac{1}{|\mathcal V|}\sum_{x\in\mathcal V}\widehat r_x(W,H),
\label{eq:validation}
\end{equation}
where $\widehat r_x$ denotes the measured outcome for task $x$, averaged over evaluation rollouts when applicable. Validation controls harness acceptance; test outcomes do not enter the optimization loop.

\subsection{Trajectory-driven alternating updates}
At round $t$, the model optimizer $\mathcal M$ performs a block of agentic RL with $H_t$ fixed:
\begin{equation}
(W_{t+1},\mathcal T_t)=\mathcal M(W_t;H_t,\mathcal D_{\mathrm{train}}).
\label{eq:weight}
\end{equation}
The trajectory batch $\mathcal T_t$ records decisions, tool interactions, and outcomes observed during training. It provides feedback on how the model operates under $H_t$, including failures that a harness revision may address. These executions can come from intermediate policies within the RL block. The updated weights are retained before harness refinement.

The harness optimizer then proposes
\begin{equation}
H'_t=\mathcal S(H_t,\mathcal T_t).
\label{eq:proposal}
\end{equation}
We instantiate $\mathcal S$ with trajectory-driven refinement adapted from HarnessEvolve~\citep{harnessevolve}. Failed executions provide diagnostic evidence for identifying recurring errors and proposing targeted edits. Reusing $\mathcal T_t$ avoids a separate collection pass for those diagnostic executions.

The harness optimizer $\mathcal S$ proposes one candidate revision per round. \method{} evaluates that candidate with $W_{t+1}$. The proposal stage determines \emph{what to change}, and the validation gate determines \emph{whether that change guides the next training round}.

\subsection{Validation-gated harness updates}

Weight updates are guided by gradients of an explicit RL objective, whereas our harness optimizer proposes discrete revisions from trajectory feedback. Such revisions do not necessarily improve task performance with the updated model. Before using a revision for subsequent training, we therefore compare the incumbent and candidate with the same checkpoint, validation tasks, and evaluator:

\begin{equation}
\begin{aligned}
b_t &= \Rval(W_{t+1},H_t), \\
c_t &= \Rval(W_{t+1},H'_t), \\
\Delta_t^H &= c_t-b_t.
\end{aligned}
\label{eq:comparison}
\end{equation}
The gate applies a strict-improvement rule:
\begin{equation}
H_{t+1}=\begin{cases}
H'_t,&\Delta_t^H>0,\\
H_t,&\Delta_t^H\leq0.
\end{cases}
\label{eq:gate}
\end{equation}
A tie retains the incumbent. Acceptance yields $(W_{t+1},H'_t)$; rejection yields $(W_{t+1},H_t)$. The retained harness governs the next round's rollouts and therefore influences subsequent weight updates. Holding $W_{t+1}$ fixed makes the comparison specific to the model that will actually use the revision.

The gate acts only on harness revisions; the updated weights are retained for the next round.
Algorithm~\ref{alg:vace} summarizes the alternating optimization loop.

\begin{algorithm}[t]
\caption{\method{}: alternating updates with validation-gated harness acceptance}
\label{alg:vace}
\small
\textbf{Input:} Initial pair $(W_0,H_0)$; training tasks $\mathcal D_{\mathrm{train}}$; validation tasks $\mathcal V$; rounds $T$; optimizers $\mathcal M,\mathcal S$.\par
\textbf{Output:} Updated model and harness $(W_T,H_T)$.\par\smallskip
\begin{tabularx}{\linewidth}{@{}r@{\quad}X@{}}
1: & \textbf{for} $t=0,\ldots,T-1$ \textbf{do} \\
2: & \quad $(W_{t+1},\mathcal T_t)\gets\mathcal M(W_t;H_t,\mathcal D_{\mathrm{train}})$ \\
3: & \quad $H'_t\gets\mathcal S(H_t,\mathcal T_t)$ \\
4: & \quad $b_t\gets\Rval(W_{t+1},H_t)$ \\
   & \quad $c_t\gets\Rval(W_{t+1},H'_t)$ \\
5: & \quad \textbf{if} $c_t>b_t$ \textbf{then} \\
6: & \qquad $H_{t+1}\gets H'_t$ \\
7: & \quad \textbf{else} \\
8: & \qquad $H_{t+1}\gets H_t$ \\
9: & \quad \textbf{end if} \\
10: & \quad Log $(b_t,c_t)$ and the gate decision \\
11: & \textbf{end for} \\
12: & \textbf{return} $(W_T,H_T)$ \\
\end{tabularx}
\end{algorithm}
\section{Experiments}
\label{sec:experiments}

\subsection{Tasks and experimental setup}
\label{sec:setup}
We study Qwen3.5-9B~\citep{qwen35} on OfficeQA~\citep{officeqa} and AutomationBench~\citep{automationbench}. OfficeQA tests document-grounded reasoning, while AutomationBench tests business workflows involving tools and workplace applications. Uni-Agent~\citep{uniagent_github} connects execution, trajectory collection, agentic RL, and harness refinement. Methods share the base model, initial harness specification, and evaluator within each dataset, while each condition collects its own rollouts.

For OfficeQA, we use a custom partition of 84 training, 53 validation, and 109 test questions. Each question receives a binary correctness score of $0$ or $1$, and we report accuracy as a percentage. For AutomationBench, we evaluate a 209-task subset from HR, Marketing, and Finance, split into 80 training, 58 validation, and 71 test tasks; Table~\ref{tab:automationbench_splits} gives the domain-level counts.

\begin{table}[htbp]
\centering\small
\caption{AutomationBench task counts by domain and split used in our experiments.}
\label{tab:automationbench_splits}
\setlength{\tabcolsep}{8pt}
\begin{tabular}{@{}lrrrr@{}}
\toprule
Split & HR & Marketing & Finance & Total \\
\midrule
Train & 26 & 27 & 27 & 80 \\
Validation & 15 & 22 & 21 & 58 \\
Test & 24 & 23 & 24 & 71 \\
\midrule
Total & 65 & 72 & 72 & 209 \\
\bottomrule
\end{tabular}
\end{table}

We use GRPO~\citep{grpo} with DAPO-style oversampling and filtering of groups whose advantages are all zero~\citep{dapo}, using only outcome rewards: binary correctness on OfficeQA and dense partial credit on AutomationBench.

For each method, we select the checkpoint and corresponding harness with the highest validation score during optimization and report the mean of three test evaluations. Figure~\ref{fig:method_comparison} in Appendix~\ref{app:selection} shows the individual scores and their sample standard deviations, which reflect repeated evaluation of one trained agent.

We first compare joint model--harness adaptation with optimizing either component alone (\textbf{RQ1}). We then compare \method{} with WHALE-style ungated alternation (\textbf{RQ2}) and SIA-style harness-first optimization (\textbf{RQ3}). Finally, we inspect individual rounds to understand how harness proposals and model updates contribute to the observed trajectory. Scores are percentages and absolute differences are percentage points (pp).

\subsection{Compared methods}
\label{sec:comparisons}
We compare \method{} with five baselines: a static agent, weight-only RL (RL-only), Harness-only, SIA-style, and WHALE-style.

The \textbf{static agent} keeps $(W_0,H_0)$ fixed. \textbf{Harness-only} fixes $W_0$ and uses the same harness optimizer and validation gate as \method{}. \textbf{RL-only} (weight-only RL) fixes $H_0$. \textbf{SIA-style}~\citep{sia} reuses the harness-only optimization phase, selects its best active harness, and then keeps that harness fixed throughout RL. \textbf{WHALE-style}~\citep{whale} uses the same optimizers and alternating update order as \method{}, but commits the output of each harness phase without the outer validation gate. \textbf{\method{}} follows Algorithm~\ref{alg:vace}.

\paragraph{Scheduling controls.}
SIA-style and WHALE-style adapt the scheduling ideas of the original methods within our implementation. They use the same weight and harness optimizers as \method{} and differ in update order or the outer validation gate. Appendix~\ref{app:controls} summarizes the optimization settings.

\subsection{Main results}
\label{sec:main_comparison}
\begin{table}[t]
\centering\small
\caption{Mean test scores (\%) from three evaluation runs of each selected model--harness pair: accuracy on OfficeQA and mean partial credit on AutomationBench. Bold marks the highest listed mean in each column.}
\label{tab:main_results}
\setlength{\tabcolsep}{4pt}
\begin{tabular}{@{}lccccc@{}}
\toprule
\multirow{2}{*}{Method} & \multirow{2}{*}{OfficeQA} & \multicolumn{4}{c}{AutomationBench} \\
\cmidrule(lr){3-6}
 & & HR & Marketing & Finance & Overall \\
\midrule
Static agent & $32.11$ & $43.86$ & $52.57$ & $54.44$ & $50.26$ \\
Harness-only & $35.28$ & $55.03$ & $64.53$ & $65.42$ & $61.62$ \\
RL-only & $38.83$ & $63.98$ & $67.55$ & $66.85$ & $66.10$ \\
SIA-style~\citep{sia} & $40.67$ & $56.36$ & $68.83$ & $71.01$ & $65.35$ \\
WHALE-style~\citep{whale} & $40.67$ & $61.50$ & $\mathbf{72.30}$ & $71.11$ & $68.25$ \\
\method{} & $\mathbf{45.26}$ & $\mathbf{75.18}$ & $72.03$ & $\mathbf{78.25}$ & $\mathbf{75.19}$ \\
\bottomrule
\end{tabular}
\end{table}

\paragraph{Joint adaptation and single-component optimization (RQ1).}
Table~\ref{tab:main_results} reports 45.26\% accuracy for \method{} on OfficeQA and 75.19\% mean partial credit on AutomationBench. These are 13.15 and 24.94 pp above the static agent, and 9.98 and 13.58 pp above harness-only optimization. \method{} exceeds RL-only by 6.43 pp on OfficeQA (45.26\% versus 38.83\%) and 9.09 pp on AutomationBench (75.19\% versus 66.10\%).

\paragraph{Validation-gated versus ungated alternation (RQ2).}
\method{} exceeds WHALE-style by 4.59 pp on OfficeQA (45.26\% versus 40.67\%) and 6.95 pp on AutomationBench (75.19\% versus 68.25\%). Both conditions update the model and harness repeatedly; their design differs in whether a harness proposal must pass the outer gate before the next RL stage. Section~\ref{sec:gate_analysis} examines the corresponding same-checkpoint acceptance decisions.

\paragraph{Repeated alternation versus a harness-first schedule (RQ3).}
On AutomationBench, \method{} exceeds SIA-style by 9.85 pp (75.19\% versus 65.35\%). On OfficeQA, the SIA-style mean is 40.67\%, compared with 45.26\% for \method{}, a difference of 4.59 pp in favor of \method{} (Table~\ref{tab:main_results}).

\subsection{Why validate harness proposals?}
\label{sec:gate_analysis}
\begin{table}[t]
\centering\small
\caption{Phase-level validation summary (\%). A/T/R counts accepted, tied, and rejected harness proposals.}
\label{tab:run_summary}
\setlength{\tabcolsep}{4pt}
\begin{tabular}{@{}lccccc@{}}
\toprule
Dataset & Initial & Final retained & Best & H trials & A/T/R \\
\midrule
OfficeQA & 30.28 & 47.17 & 60.37 & 12 & 7/2/3 \\
AutomationBench & 53.08 & 81.42 & 81.42 & 32 & 18/0/14 \\
\bottomrule
\end{tabular}
\end{table}

The phase-level trajectories contain 44 harness comparisons: 25 accepted proposals, 17 regressions, and 2 ties (Table~\ref{tab:run_summary}). OfficeQA accepts 7 of 12 proposals, while AutomationBench accepts 18 of 32. Many proposals fail to improve validation performance at the checkpoint where they are evaluated.
\begin{table}[t]
\centering\small
\caption{Same-checkpoint examples from the phase-level trajectories: the first accepted and first rejected proposal in each dataset. Scores are percentages; changes are percentage points.}
\label{tab:harness_examples}
\setlength{\tabcolsep}{4pt}
\begin{tabular}{@{}lrccrc@{}}
\toprule
Dataset & RL update & Incumbent & Candidate & Change & Decision \\
\midrule
OfficeQA & 5 & 41.51 & 45.28 & $+3.77$ & Accept \\
OfficeQA & 30 & 60.37 & 49.06 & $-11.31$ & Reject \\
AutomationBench & 5 & 53.81 & 61.07 & $+7.26$ & Accept \\
AutomationBench & 15 & 70.93 & 64.50 & $-6.43$ & Reject \\
\bottomrule
\end{tabular}
\end{table}

Table~\ref{tab:harness_examples} shows the first accepted and first rejected proposal in each trajectory. Proposals produce both improvements and regressions despite being generated to address observed failures. The gate retains the incumbent when the proposed revision lowers validation performance.

\subsection{Alternating optimization dynamics}
\label{sec:dynamics}
\begin{figure}[t]
\centering
\includegraphics[width=\linewidth]{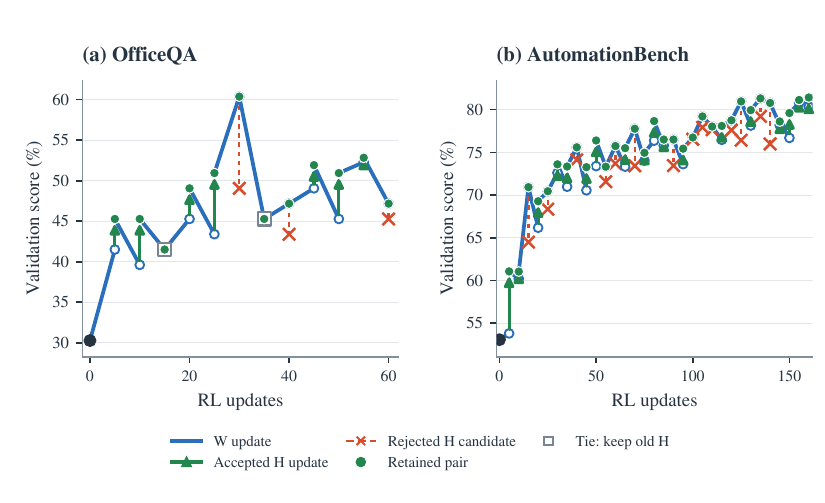}
\caption{Phase-level validation dynamics. Blue segments connect a retained pair to the next post-weight evaluation; green arrows mark accepted harness updates, red crosses rejected candidates, and outlined squares ties. Green dots identify retained pairs, from which the next weight stage starts.}
\label{fig:coevolution_dynamics}
\end{figure}

Figure~\ref{fig:coevolution_dynamics} shows the alternating validation trajectories on both datasets. The retained validation score on OfficeQA rises from 30.28\% to 47.17\%, a change of 16.89 pp. On AutomationBench, it rises from 53.08\% to 81.42\%, a change of 28.34 pp. Progress is not monotonic: weight stages can reduce the score, and the gate only compares harnesses at the current updated checkpoint.

The highest validation scores are 60.37\% on OfficeQA and 81.42\% on AutomationBench. Appendix~\ref{app:selection} provides the corresponding validation curves.
\section{Discussion and limitations}
\label{sec:discussion}

\paragraph{What the co-evolution loop contributes.}
\method{} connects two uses of execution experience: updating model behavior through RL and identifying harness revisions from the collected trajectories. Validation then determines which revision accompanies the updated model into the next round. In our experiments, \method{} achieves higher mean test scores than RL-only and WHALE-style on both datasets. The phase-level records contain both beneficial and harmful harness proposals, supporting the use of an explicit acceptance decision before subsequent training. Cross-evaluating checkpoints and harness versions could further test how their compatibility changes during training.

\paragraph{Validation reuse and evaluation noise.}
Repeated use of the validation set for harness acceptance can introduce adaptive selection effects~\citep{dwork2015}. The strict-improvement gate also relies on noisy point estimates.

\paragraph{Scope and computational cost.}
The study covers one model size and two agent benchmarks, with harness edits focused on skills and execution guidance. Broader environments and model scales remain to be explored. Harness refinement and validation add computational overhead to model training.

\paragraph{Fixed scheduling.}
\method{} uses a fixed alternating schedule. Feedback-driven scheduling could choose when to update each component based on task outcomes and training progress.

\Needspace{10\baselineskip}
\section{Conclusion}
We presented \method{}, an agentic RL framework that alternates model training with trajectory-driven harness refinement. A validation comparison at the updated checkpoint determines which harness guides the next training stage. In our experiments on OfficeQA and AutomationBench, \method{} achieves higher mean test scores than weight-only RL and WHALE-style ungated alternation. The optimization records also show that proposed harness revisions can reduce validation performance, motivating an explicit acceptance decision before further training. Future work will evaluate VACE on more datasets and explore feedback-driven scheduling across additional models and environments.
\clearpage
\begingroup\small\raggedright

\endgroup
\clearpage
\appendix
\section{Optimization settings}
\label{app:controls}
Table~\ref{tab:budgets} summarizes the optimization settings for the compared methods.

\begin{table}[H]
\centering\small
\caption{Optimization settings. $B_W$ and $B_H$ denote the planned RL and harness-proposal caps.}
\label{tab:budgets}
\setlength{\tabcolsep}{5pt}
\begin{tabular}{@{}lccc@{}}
\toprule
Method & RL cap & H cap & Harness gate \\
\midrule
Static agent & 0 & 0 & --- \\
Harness-only & 0 & $B_H$ & Yes \\
RL-only & $B_W$ & 0 & --- \\
SIA-style & $B_W$ & $B_H$ & Yes \\
WHALE-style & $B_W$ & $B_H$ & No \\
\method{} & $B_W$ & $B_H$ & Yes \\
\bottomrule
\end{tabular}
\end{table}

\paragraph{Single-component conditions.}
Harness-only optimization fixes $W_0$ and collects diagnostic trajectories under the current harness. It uses the same harness optimizer and validation gate as \method{}. Weight-only RL fixes $H_0$ and applies the common weight optimizer throughout.

\paragraph{WHALE-style.}
This condition uses the same weight and harness optimizers, editable scope, and initial-state specification as \method{}. It commits the harness phase output without the outer validation gate.

\paragraph{SIA-style.}
SIA-style reuses the harness-only optimization phase before weight training. In this condition, harness candidates are checked with the validation gate described in Section~\ref{sec:method}; the best active harness is selected and remains fixed during RL.

\paragraph{Separate training runs.}
Each weight-training condition follows its own trajectory; SIA-style reuses the Harness-only phase before weight training.

\paragraph{Paired evaluation.}
The incumbent and candidate are evaluated on identical task IDs at the same checkpoint using the same evaluator.
\section{Additional experimental results}
\label{app:selection}
\begin{figure}[H]
\centering
\includegraphics[width=\linewidth]{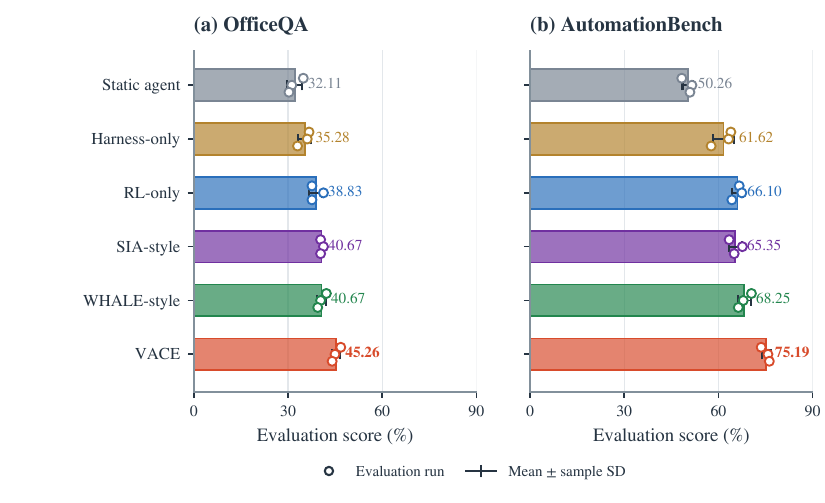}
\caption{Six-condition score comparison. Bars show means; points show three test evaluations of each selected model--harness pair, and whiskers show their sample standard deviations.}
\label{fig:method_comparison}
\end{figure}

\begin{figure}[H]
\centering
\includegraphics[width=\linewidth]{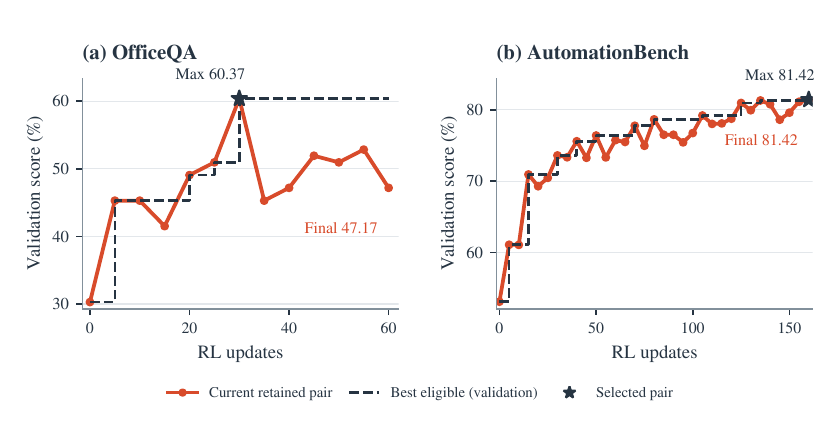}
\caption{Validation trajectories. Solid curves show the scores after each round, dashed curves show the best scores reached so far, and stars mark the maxima.}
\label{fig:checkpoint_selection}
\end{figure}
\end{document}